\documentclass[12pt]{article}

\usepackage{graphicx}
\usepackage{amssymb}
\usepackage{JASA_manu}
\usepackage{hyperref}

\usepackage{fancyvrb}
\usepackage{url}
\usepackage{algorithm}
\usepackage{algorithmic}
\usepackage{amssymb,amsmath}
\usepackage{graphicx}
\usepackage{psfrag,subfigure} 
\usepackage{color}
\usepackage{wrapfig}
\usepackage{lipsum}
\usepackage{bm}
\usepackage{times}
\usepackage{wrapfig}

\definecolor{violet}{RGB}{143, 0, 255}

\newcommand{\xxx}{STRADS}

\newcommand{\bbeta}{\boldsymbol{\beta}}

\newcommand{\bA}{\mathbf{A}}
\newcommand{\bW}{\mathbf{W}}
\newcommand{\bH}{\mathbf{H}}

\newcommand{\bX}{\mathbf{X}}

\newcommand{\bh}{\mathbf{h}}

\newcommand{\by}{\mathbf{y}}
\newcommand{\bx}{\mathbf{x}}

\newcommand{\bw}{\mathbf{w}}

\begin{document}

\title{Primitives for Dynamic Big Model Parallelism}
\author{Seunghak Lee,
Jin Kyu Kim, 
Xun Zheng,
Qirong Ho, 
Garth A. Gibson, Eric P. Xing$^{*}$ \\ 
School of Computer Science \\ 
Carnegie Mellon University, Pittsburgh, PA, U.S.A. \\ 
$^{*}$email: \texttt{epxing@cs.cmu.edu} }

\maketitle

\begin{center}
\textbf{Abstract}
\end{center}
When training large machine learning models with many variables or parameters,
a single machine is often inadequate since the model
may be too large to fit in memory, while training
can take a long time even with stochastic updates.
A natural recourse is to turn to distributed cluster computing,
in order to harness additional memory and processors.
However, naive, unstructured parallelization of ML algorithms can make
inefficient use of distributed memory, while failing to obtain
proportional convergence speedups --- or can even result in divergence.
We develop a framework of primitives for dynamic model-parallelism, STRADS, in order
to explore partitioning and update scheduling of
model variables in distributed ML algorithms ---
thus improving their memory efficiency
while presenting new opportunities to speed up convergence without
compromising inference correctness. We demonstrate the efficacy of model-parallel
algorithms implemented in STRADS versus popular implementations
for Topic Modeling, Matrix Factorization and Lasso.

\vspace*{.3in}


\section{Introduction}
Sensory techniques and digital storage media have improved at a breakneck pace, leading to massive ``Big Data" collections that have been the focus of recent efforts to achieve scalable machine learning (ML). 
Numerous {\it data-parallel} algorithmic and system solutions, both heuristic and principled,
have been proposed to speed up inference on Big Data~\cite{dean2008mapreduce,low2012distributed,malewicz2010pregel,zaharia2010spark};
however, large-scale ML also encompasses {\it Big Model} problems \cite{fan2009ultrahigh}, 
in which models with millions if not billions of variables and/or parameters 
(such as in deep networks \cite{dean2012large} or large-scale topic models \cite{newman2009distributed}) 
must be estimated from big (or even modestly-sized) datasets. 
These Big Model problems seem to have received less attention in ML communities, which,
in turn, has limited their application to real-world problems.

Big Model problems are challenging because
a large number of model variables must be efficiently updated until model convergence.
Data-parallel algorithms such as stochastic gradient descent~\cite{zinkevich2009slow}
concurrently update all model variables given a subset of data samples, but this requires
every worker to have full access to all global variables --- which can be very large,
such as the billions of variables in Deep Neural Networks~\cite{dean2012large},
or this paper's large scale topic model with 22M bigrams by 10K topics (200 {\it billion} variables) and
matrix factorization with rank 2K on a 480K-by-10K matrix (1B variables).
Furthermore, data-parallelism does not consider the possibility
that some variables may be more important than others for algorithm convergence,
a point that we shall demonstrate through our Lasso implementation (run on 100M coefficients).
On the other hand, model-parallel algorithms such as coordinate descent~\cite{bradley2011parallel}
are well-suited to Big Model problems, because parallel workers focus on subsets of model variables.
This allows the variable space to be partitioned for memory efficiency, and also allows
some variables to be prioritized over others.
However, model-parallel algorithms are usually developed for a specific application 
such as Matrix Factorization~\cite{gemulla2011large} or Lasso~\cite{bradley2011parallel} ---
thus, there is utility in developing programming primitives that can tackle the common
challenges of Big Model problems, while also exposing new opportunities such as variable prioritization.

Existing distributed frameworks such as MapReduce~\cite{dean2008mapreduce} and GraphLab~\cite{low2012distributed}
have shown that common primitives such as Map/Reduce or Gather/Apply/Scatter 
can be applied to a variety of ML applications. Crucially,
these frameworks automatically decide which variable to update next --- MapReduce executes
all Mappers at the same time, followed by all Reducers, while GraphLab chooses the next node
based on its ``chromatic engine" and the user's choice of graph consistency model.
While such {\it automatic scheduling} is convenient, it does not offer the fine-grained control needed
to avoid parallelization of variables with subtle interdependencies not seen in the superficial
problem or graph structure
(which can then lead to algorithm divergence, as in Lasso~\cite{bradley2011parallel}). Moreover,
it does not allow users to explicitly prioritize variables based on new criteria.

\renewcommand{\arraystretch}{1.2}
\setlength{\tabcolsep}{3pt}\begin{table}
\label{tab:app_summary}
\centering
{\scriptsize  
\caption{\small Summary of LDA, MF, and Lasso on STRADS (detailed pseudocode is in the relevant sections).}
    \vspace{.8em}
    \begin{tabular}{|l|c|c|c|c|}
    \hline
     & {\bf Schedule} & {\bf Push} and {\bf Pull} & \bf{Largest \xxx{} experiment} \\ \hline
    Topic Modeling (LDA) & Word rotation scheduling & Collapsed Gibbs sampling & 10K topics, 3.9M docs with 21.8M vocab  \\ \hline    
    MF  & Round-robin scheduling & Coordinate descent& rank-2K, 480K-by-10K matrix \\ \hline
    Lasso & Dynamic priority scheduling & Coordinate descent & 100M features, 50K samples \\ \hline 
    \end{tabular}
}
\end{table} 

To improve upon these frameworks, we develop new primitives for dynamic Big Model parallelism:
{\bf schedule}, {\bf push} and {\bf pull},
which are executed by our STRADS system (STRucture-Aware Dynamic Scheduler).
These primitives are inspired by the simplicity and wide applicability of MapReduce,
but also provide the fine control needed to explore novel ways of performing dynamic model-parallelism.
{\bf Schedule} specifies the next subset of model variables to be updated in parallel,
{\bf push} specifies how individual workers compute partial results on those variables,
and {\bf pull} specifies how those partial results are aggregated to perform the full variable update.
A final ``automatic primitive", {\bf sync}, ensures that distributed workers have up-to-date
values of the model variables, and is automatically executed at the end of {\bf pull}; the user does
not need to implement {\bf sync}.
To explore the utility of STRADS, we implement {\bf schedule}, {\bf push} and {\bf pull}
for three popular ML applications (Table \ref{tab:app_summary}): Topic Modeling (LDA),
Lasso, and Matrix Factorization (MF). Our goal is not to best specialized implementations in performance,
but to demonstrate that STRADS primitives enable Big Model problems to be solved with modest programming effort.
In particular, we tackle
topic modeling with 3.9M docs, 10K topics and 21.8M vocabulary ($200$B variables),
MF with rank-2K on a 480K-by-10K matrix ($1$B variables), and Lasso with 100M features (100M variables).

\section{Primitives for Dynamic Model Parallelism}

\begin{figure}[t]
\centering
\includegraphics[width=1\textwidth]{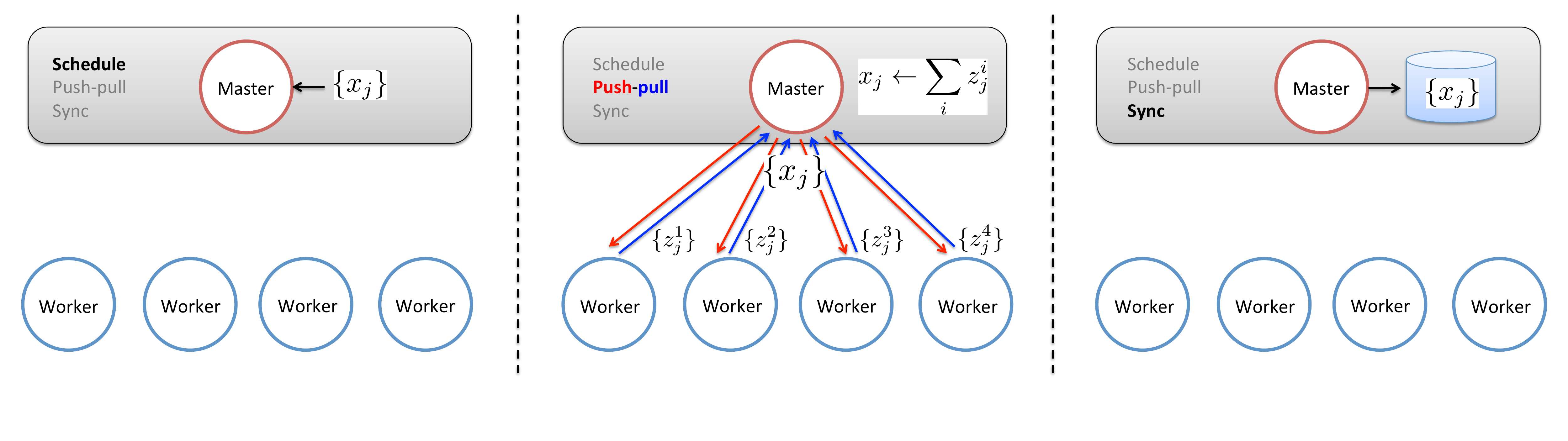}
\vspace{-1.2cm}
\caption{
{\small High-level view of our STRADS primitives for dynamic model parallelism.}
}  
\label{fig:premitives}
\end{figure}

\begin{figure}[t]
{\scriptsize
\begin{Verbatim}[frame=single]
// Generic STRADS application
\end{Verbatim}
\vspace{-0.3cm}
\begin{Verbatim}[frame=single]
schedule() {
  // Select U vars x[j] to be sent
  // to the workers for updating
  ...
  return (x[j_1], ..., x[j_U])
}
\end{Verbatim}
\vspace{-0.3cm}
\begin{Verbatim}[frame=single]
push(worker = p, vars = (x[j_1],...,x[j_U])) {
  // Compute partial update z for U vars x[j]
  // at worker p
  ...
  return z
}
\end{Verbatim}
\vspace{-0.3cm}
\begin{Verbatim}[frame=single]
pull(workers = [p], vars = (x[j_1],...,x[j_U]),
     updates = [z]) {
  // Use partial updates z from workers p to
  // update U vars x[j]. sync() is automatic.
  ...
}
\end{Verbatim}
}
\caption{\small {\bf STRADS user-defined primitives:} {\bf schedule}, {\bf push}, {\bf pull}.
We show the basic functional signature of each primitive, using pseudocode.}
\label{fig:strads_sample}
\end{figure}

``Model parallelism" refers to parallelization of an ML algorithm over the space of shared model variables, rather than the space of (usually i.i.d.) data samples. At a high level, model variables are the changing intermediate quantities that an ML algorithm iteratively updates, until convergence is reached. For example, the coefficients in regression are model variables, which are iteratively updated using algorithmic strategies like coordinate descent.

\begin{figure}[t]
\centering
\includegraphics[width=0.5\textwidth]{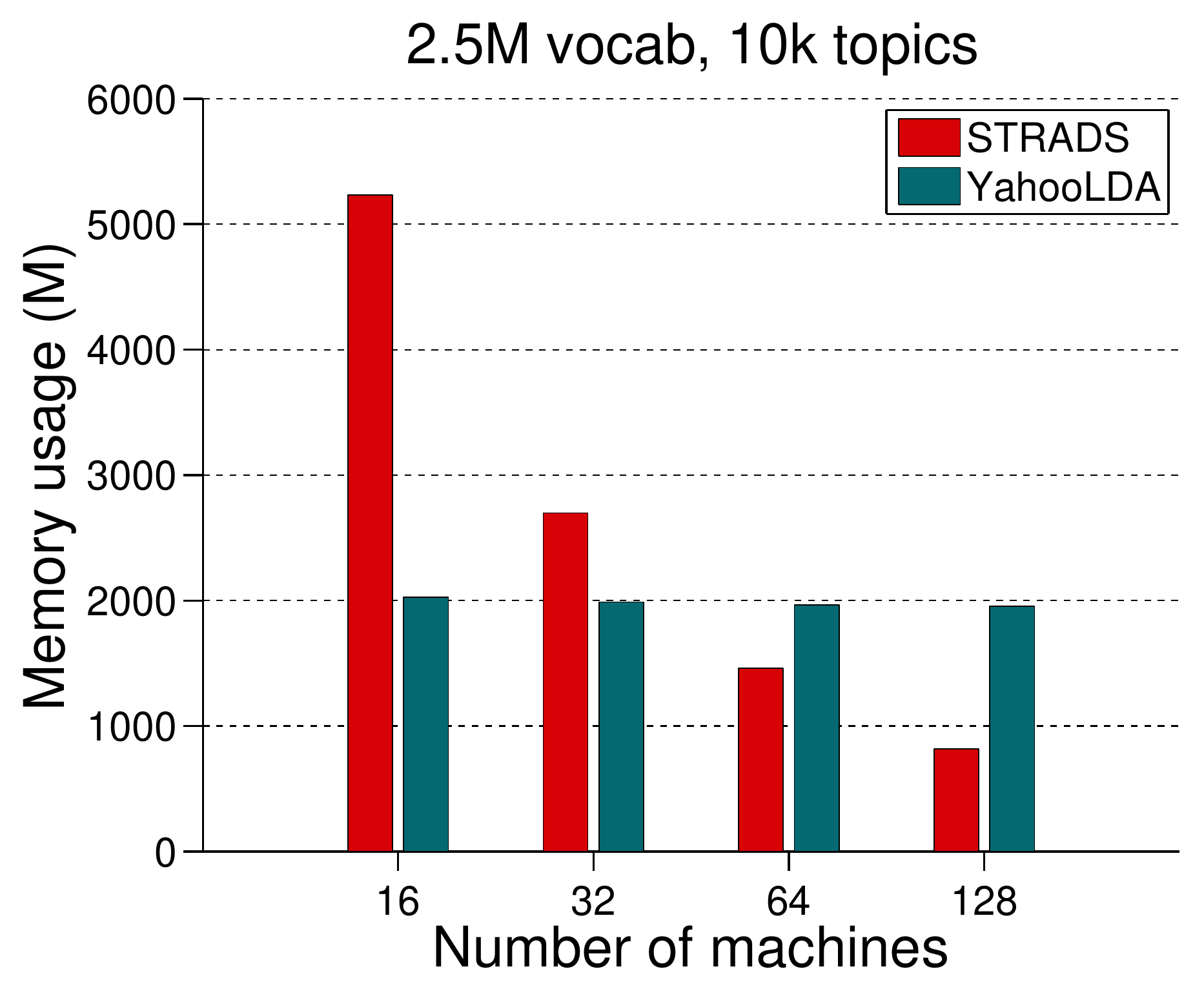}
\caption{\small
{\bf Topic Modeling: Memory usage per machine}, for model-parallellism (STRADS) vs data-parallellism (YahooLDA).
With more machines, STRADS LDA uses {\it less memory per machine}, because it explicitly partitions
the model space.
}
\label{fig:lda_memoryusage_permachine}
\end{figure}

Model parallelism can be contrasted with data parallelism, in which the ML algorithm is parallelized over individual data samples, such as in stochastic optimization algorithms~\cite{zinkevich2010parallelized}. A key advantage of the model-parallel approach is that it explicitly partitions the model variables into subsets, allowing ML problems with massive model spaces to be tackled on machines with limited memory. Figure~\ref{fig:lda_memoryusage_permachine} shows this advantage: for topic modeling, STRADS uses less memory per machine as the number of machines increases, unlike the data-parallel YahooLDA algorithm. As our experiments will confirm, this means that STRADS can handle larger ML models (given sufficient machines), whereas YahooLDA is strictly constrained by the memory of the smallest machine. This has practical consequences --- STRADS LDA can handle bigram vocabularies with over 20 million term-pairs on modest hardware (enabling large-scale topic modeling applications), while YahooLDA cannot.

To enable users to systematically and programmatically exploit model parallelism, our proposed STRADS framework defines a set of primitives. Similar to the map-reduce paradigm, these primitives are functions that a user writes
for his/her ML problem, and STRADS repeatedly executes these functions to create an iterative model-parallel algorithm
(Figures~\ref{fig:premitives}, \ref{fig:strads_sample}). Our primitives are {\bf schedule}, {\bf push} and {\bf pull},
and a single ``round" or iteration of STRADS executes them in that order. In addition, there is an automatic primitive, {\bf sync}, which the user does not have to write.

\paragraph{Schedule:}
This primitive determines the parallel order for updating model variables; as
shown in Figure~\ref{fig:strads_sample}, {\bf schedule} selects
$U$ model variables to be dispatched for updates (Figure~\ref{fig:premitives}). Within the
{\bf schedule} function, the programmer may access all data $D$ and all model variables $x$,
in order to decide which $U$ variables to dispatch.
The simplest possible {\bf schedule} is to select model variables according to a fixed sequence,
or drawn uniformly at random. As we shall later see, {\bf schedule} also allows
model variables to be selected in a way that:
(1) dynamically focuses workers on the fastest-converging variables, while avoiding already-converged variables;
(2) avoids parallel dispatch of variables with inter-dependencies, which can lead to divergence and incorrect execution.

\paragraph{Push and Pull:}
These primitives control the flow of model variables $x$ and data $D$ from the master scheduler machines(s) to and from the workers (Figure~\ref{fig:premitives}). The {\bf push} primitive dispatches a set of variables $\{x_{j_1},\dots,x_{j_U}\}$ to each worker $p$, which then computes a partial update $z$ for $\{x_{j_1},\dots,x_{j_U}\}$ (or a subset of it). When writing {\bf push}, the user can take advantage of data partitioning: e.g., when only a fraction $\frac{1}{P}$ of the data samples are stored at each worker, the $p$-th worker should compute partial results $z_j^p = \sum_{D_i} f_{x_j}(D_i)$ by iterating over its $\frac{1}{L}$ data points $D_i$. The {\bf pull} primitive is used to aggregate the partial results $\{z_j^p\}$ from all workers, and commit them to the variables $\{x_{j_1},\dots,x_{j_U}\}$. Our STRADS LDA, Lasso and MF applications partition the data samples uniformly over machines.

\paragraph{Synchronization:}
The model variables $x$ are globally accessible through a distributed, partitioned key-value store
(represented by standard arrays in our pseudocode).
{\bf Sync} is a built-in primitive that ensures all {\bf push} workers can access up-to-date
model variables, and is automatically executed whenever {\bf pull} writes to any variable $x[j]$.
The user does not need to implement {\bf sync}.
A variety of key-value store synchronization schemes exist, such as
Bulk Synchronous Parallel (BSP), Stale Synchronous Parallel (SSP)~\cite{ho2013more}, and
Asynchronous Parallel (AP).
Each presents a different trade-off:
BSP is simple and correct but easily bottlenecked by slow workers,
AP is usually effective but risks algorithmic errors and divergence because
it has no error guarantees, and SSP is fast and guaranteed to converge
but requires more engineering work and parameter tuning.
In this paper, we use BSP for {\bf sync} throughout;
we leave the use of alternative schemes like SSP or AP as future work.

\section{Harnessing Model-Parallelism in ML Applications through STRADS}\label{sec:app}

In this section, we shall explore how users can apply model-parallelism
to their own ML applications, using the STRADS primitives. We shall cover
3 ML application case studies, with the intent of showing that model-parallelism
in STRADS can be simple and effective, yet also powerful enough to expose new and interesting
opportunities for speeding up distributed ML.

\subsection{Latent Dirichlet Allocation (LDA)}
\label{sec:lda_description}

We introduce STRADS programming through topic modeling via LDA~\cite{blei2003latent}.
Big LDA models provide a strong use case for model-parallelism: when thousands of topics and millions of words are used,
the LDA model contains billions of global variables, and data-parallel implementations face the difficult challenge
of providing access to all these variables; in contrast, model-parallellism explicitly divides up the variables, so that
workers only need to access a fraction at a given time.

Formally, LDA takes a corpus of $N$ documents
as input, and outputs $K$ topics (each topic is just 
a categorical distribution over all $V$ unique words in the corpus)
as well as $N$ $K$-dimensional topic vectors (soft assignments of topics to documents).
The LDA model is
\\[-0.5cm]
\begin{align*}
\mathrm{P}(\bm{W} \mid \bm{Z}, \bm{\theta}, \bm{\beta}) = \prod_{i=1}^{N} \prod_{j=1}^{M_i}
    \mathrm{P}(w_{ij} \mid z_{ij}, \bm{\beta})\mathrm{P}(z_{ij} \mid \bm{\theta}_i),
\end{align*}\\[-0.3cm]
where (1) $w_{ij}$ is the $j$-th token (word position) in the $i$-th document, (2) $M_i$ is the number
of tokens in document $i$, (3) $z_{ij}$ is the
topic assignment for $w_{ij}$, (4) $\bm{\theta}_i$ is the topic vector for document $i$, and (5) 
$\bm{\beta}$ is a matrix representing the $K$ $V$-dimensional topics.
LDA is commonly reformulated as a ``collapsed" model~\cite{griffiths2004finding}
in which $\bm{\theta},\bm{\beta}$ are integrated out for faster inference. Inference is performed using Gibbs sampling,
where each $z_{ij}$ is sampled in turn according to its distribution conditioned on all other variables,
$\mathrm{P}(z_{ij} \mid \bm{W}, \bm{Z}_{-ij})$. To perform this computation 
without having to iterate over all $\bm{W},\bm{Z}$,
sufficient statistics are kept in the form of a ``doc-topic" table $\bm{D}$ (analogous to $\bm{\theta}$),
and a ``word-topic" table $\bm{B}$ (analogous to $\bm{\beta}$). More precisely,
$D_{ik}$ counts the number of assignments $z_{ij} = k$ in doc $i$,
while $B_{vk}$ counts the number of tokens $w_{ij} = v$ such that $z_{ij} = k$.

\paragraph{STRADS implementation:}

\begin{figure}[t] 
{\scriptsize
\begin{Verbatim}[frame=single]
// STRADS LDA
\end{Verbatim}
\vspace{-0.3cm}
\begin{Verbatim}[frame=single]
schedule() {
  dispatch = []  // Empty list
  for a=1..U     // Rotation scheduling
    idx = ((a+C-1) mod U) + 1
    dispatch.append( V[q_idx] )
  return dispatch
}
\end{Verbatim}
\vspace{-0.3cm}
\begin{Verbatim}[frame=single]
push(worker = p, vars = [V_a, ..., V_U]) {
  t = []               // Empty list
  for (i,j) in W[q_p]  // Fast Gibbs sampling
    if w[i,j] in V_p
      t.append( (i,j,f_1(i,j,D,B)) )
  return t
}
\end{Verbatim}
\vspace{-0.3cm}
\begin{Verbatim}[frame=single]
pull(workers = [p], vars = [V_a, ..., V_U],
     updates = [t]) {
  for all (i,j)    // Update sufficient stats
  (D,B) = f_2([t])
}
\end{Verbatim}
}
\vspace{-0.5cm}
\caption{\small{\bf STRADS LDA pseudocode.} Definitions for $f_1,f_2,q_p$
are in the text. {\tt C} is a global model variable.}
\label{fig:strads_lda}
\end{figure}

In order to perform model-parallelism, we first identify the model variables,
and create a {\bf schedule} strategy over them. In LDA, the assignments $z_{ij}$ are the
model variables, while $\bm{D},\bm{B}$ are summary statistics over the $z_{ij}$
that are used to speed up the sampler. Our {\bf schedule} strategy equally divides the
$V$ words into $U$ subsets $V_1,\dots,V_U$ (where $U$ is the number of workers).
Each worker will only process words from one subset $V_a$ at a time.
Subsequent invocations of {\bf schedule} will ``rotate" subsets amongst workers, so that
every worker touches all $U$ subsets every $U$ invocations.
For data partitioning, we divide the document tokens $\bm{W}$ evenly across workers, and denote worker $p$'s set
of tokens by $\bm{W}_{q_p}$.

During {\bf push}, suppose that worker $p$ is assigned to subset $V_a$ by {\bf schedule}. This worker
will only Gibbs sample the topic assignments $z_{ij}$ such that (1) $(i,j) \in \bm{W}_{q_p}$ and (2) $w_{ij} \in V_a$.
In other words, $w_{ij}$ must be assigned to worker $p$, and must also be a word in $V_a$.
The latter condition is the source of model-parallelism: observe how the assignments $z_{ij}$ are chosen for sampling
based on word divisions $V_a$. Note that all $z_{ij}$ will be sampled exactly once after $U$ invocations of
{\bf schedule}.
We use the fast Gibbs sampler from~\cite{yao2009efficient} to {\bf push}
update $z_{ij} \leftarrow f_1(i,j,\bm{D},\bm{B})$,
where $f_1(\cdot)$ represents the fast Gibbs sampler equation.
The {\bf pull} step simply updates the sufficient statistics $\bm{D},\bm{B}$ using the new $z_{ij}$, and
we represent this procedure as a function $(\bm{D},\bm{B}) \leftarrow f_2([z_{ij}])$.
Figure~\ref{fig:strads_lda} provides pseudocode for STRADS LDA.

\paragraph{Model parallelism results in low error:}

\begin{figure}[t] 
\centering
\includegraphics[width=0.5\textwidth]{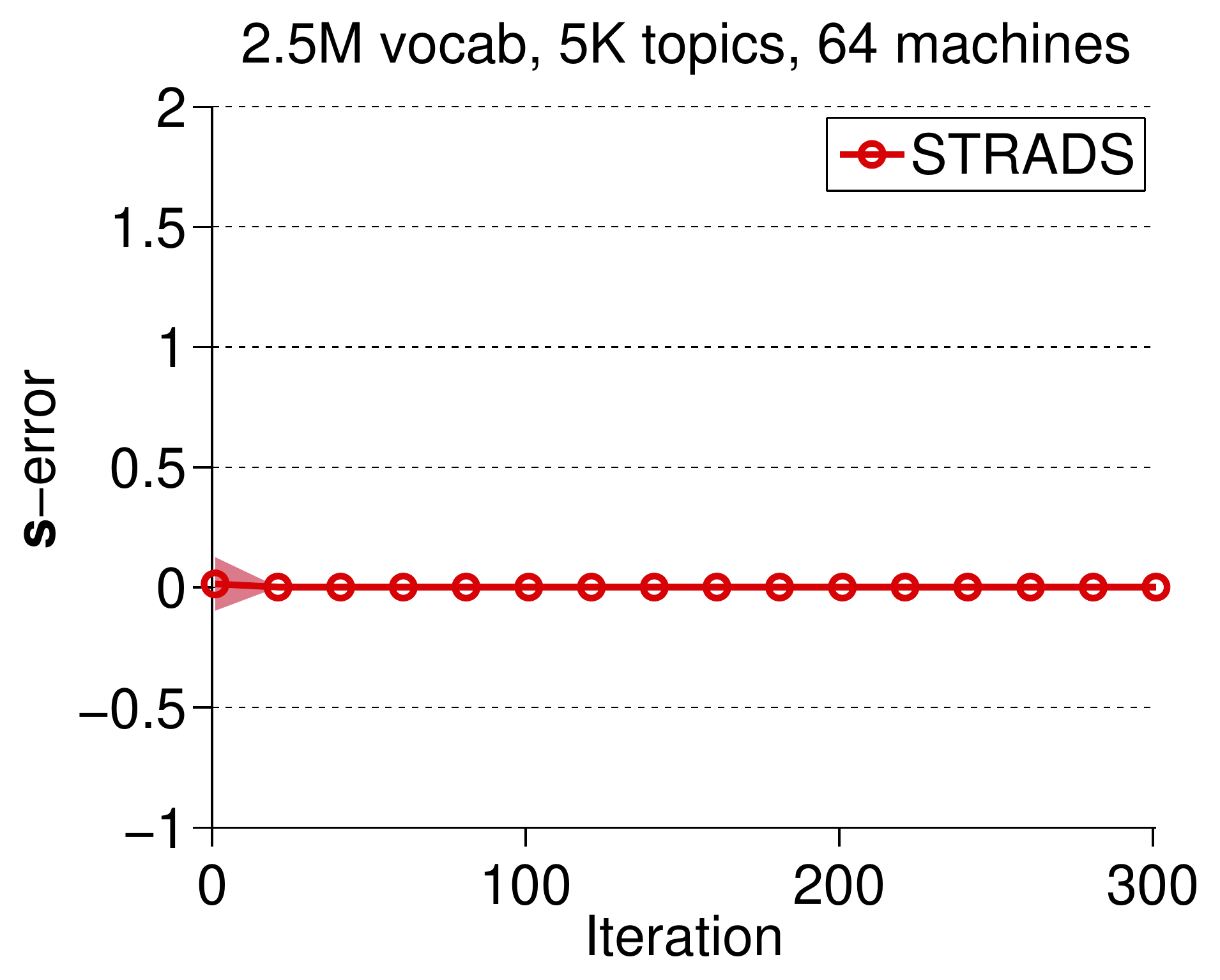}
\caption{\small
{\bf STRADS LDA:} $\bm{s}$-error $\Delta_{r,t}$
at each iteration, on the Wikipedia unigram dataset with $K=5000$ and 64 machines. 
}
\label{fig:lda_S_error}
\end{figure}

Parallel Gibbs sampling is not generally guaranteed to converge~\cite{gonzalez2011parallel},
unless the variables being parallel-sampled are conditionally independent of each other.
Because STRADS LDA assigns workers to disjoint words $V$ and documents $w_{ij}$, each worker's variables $z_{ij}$
are (almost) conditionally independent of other workers, except for a single shared dependency:
the column sums of $\bm{B}$ (denoted by $\bm{s}$, and stored as an extra row appended to $\bm{B}$),
which are required for correct normalization of the Gibbs sampler conditional distributions in $f_1()$.
The column sums $\bm{s}$ are {\bf synced} at the end of every {\bf pull}, but will go out-of-sync
during worker {\bf pushes}.
To understand how error in $\bm{s}$ affects sampler convergence, consider
the Gibbs sampling conditional distribution for a topic indicator $z_{ij}$:
\begin{align*}
\mathrm{P}(z_{ij} \mid \bm{W},\bm{Z}_{-ij})
    &\propto \mathrm{P}(w_{ij} \mid z_{ij}, \bm{W}_{-ij},\bm{Z}_{-ij}) \mathrm{P}(z_{ij} \mid \bm{Z}_{-ij}) \\
&=\frac{\gamma + B_{w_{ij},z_{ij}}}{V\gamma + \sum_{v=1}^{V} B_{v,z_{ij}}} \times
    \frac{\alpha + D_{i,z_{ij}}}{K\alpha + \sum_{k=1}^{K} D_{i,k}}.
\end{align*}
In the first term, the denominator quantity $\sum_{v=1}^{V} B_{v,z_{ij}}$ is exactly the
sum over the $z_{ij}$-th column of $\bm{B}$, i.e. $\bm{s}_{z_{ij}}$. Thus,
errors in $\bm{s}$ induce errors in the probability distribution
$U_{w_{ij}} \sim \mathrm{P}(w_{ij} \mid z_{ij}, \bm{W}_{-ij}, \bm{Z}_{-ij})$, which is just the discrete
probability that topic $z_{ij}$ will generate word $w_{ij}$. As a proxy for the error in $U$,
we can measure the difference between the true $\bm{s}$ and its local copy $\tilde{\bm{s}}^p$
on worker $p$. If $\bm{s} = \tilde{\bm{s}}^p$, then $U$ has zero error.

We can show that the error in $\bm{s}$ is empirically negligible (and hence
the error in $U$ is also small). Consider a single STRADS LDA iteration $t$,
and define its $\bm{s}$-error to be
\\[-0.3cm]
\begin{equation}
\label{eq:s_error}
\textstyle
\Delta_{t} = \frac{1}{PM} \sum_{p=1}^{P} \Vert \tilde{\bm{s}}^p - \bm{s} \Vert_1,
\end{equation}\\[-0.3cm]
where $M$ is the total number of tokens $w_{ij}$.
The $\bm{s}$-error $\Delta_{r,t}$ must lie in $[0,2]$, where 0 means no error.
Figure \ref{fig:lda_S_error} plots the $\bm{s}$-error for the ``Wikipedia unigram" dataset (refer
to our experiments section for details), for $K=5000$ topics and 64  machines (128 processor cores total).
The $\bm{s}$-error is $\le 0.002$ throughout, confirming that STRADS LDA exhibits very small parallelization error.

\subsection{Matrix Factorization (MF)}
STRADS's model-parallelism benefits other models as well: we now
consider Matrix Factorization (collaborative filtering), which can be used to
predict users' unknown preferences, given their known
preferences and the preferences of others. While most MF implementations tend to focus on small
decompositions with rank $K\approx100$~\cite{zhou2008large,gemulla2011large,yu2012scalable},
we are interested in enabling larger decompositions with rank $>1000$, where the
much larger factors (billions of variables) pose a challenge for purely data-parallel algorithms
(such as naive SGD) that need to share all variables across all workers;
again, STRADS addresses this by explicitly dividing variables across workers.

Formally, MF takes an incomplete matrix $\bA \in \mathbb{R}^{N\times{}M}$ as input,
where $N$ is the number of users, and $M$ is the number of items/preferences.
The idea is to discover  rank-$K$ matrices 
$\bW \in \mathbb{R}^{N\times{}K}$ and $\bH\in \mathbb{R}^{K\times{}M}$
such that $\bW \bH \approx \bA$. Thus, the product $\bW\bH$ can be used
to predict the missing entries (user preferences).
Formally, let $\Omega$ be the set of indices of observed entries in $\bA$,
let $\Omega^i$ be the set of observed column indices in the $i$-th row of $\bA$, and
let $\Omega_j$ be the set of observed row indices in the $j$-th column of $\bA$.
Then, the MF task is defined as an optimization problem:
\\[-0.3cm]
\begin{equation}
\textstyle
\min_{\bW,\bH} \sum_{(i,j) \in \Omega} (a_{j}^i - \bw^i\bh_j)^2 + \lambda ( \left\| \bW \right\|_F^2 + \left\| \bH \right\|_F^2).
\label{eq:mf}
\end{equation}\\[-0.3cm]
This can be solved using parallel CD \cite{yu2012scalable}, with the following update rule for $\bH$:
\\[-0.4cm]
\begin{align}
\label{eq:mf_update}
(h_j^k)^{(t)} & \leftarrow \frac{\sum_{i \in \Omega_j}\left\{r_j^i+(w_k^i)^{(t-1)}  (h_j^k)^{(t-1)}\right\}(w_k^i)^{(t-1)}}
{\lambda + \sum_{i\in \Omega_j} \left\{(w_k^i)^{(t-1)}\right\}^2},
\end{align}\\[-0.3cm]
where $r_j^i = a_j^i - (\bw^i)^{(t-1)} (\bh_j)^{(t-1)}$ for all $(i,j) \in \Omega$, and a similar rule holds for $\bW$.

\paragraph{STRADS implementation:}

Our {\bf schedule} strategy is to partition the rows of $\bA$ into
$U$ disjoint index sets $q_p$, and the columns of $\bA$ into $U$ disjoint index sets $r_p$.
We then dispatch the model variables $\bW,\bH$ in round-robin fashion, according to these sets $q_p,r_p$.
To update elements of $\bW$, each worker $p$ computes partial updates on its assigned columns $r_p$ of $\bA$ and $\bH$,
and analogously for $\bH$ and rows $q_p$ of $\bA$ and $\bW$.
The sets $q_p,r_p$ also tie neatly into data partitioning: we merely have to
divide $\bA$ into $U$ pairs of submatrices (where $U$ is the number of workers),
and store the the submatrices $\bA^{q_p}$ and $\bA_{r_p}$ at the $p$-th worker.

Consider the {\bf push} update for $\bH$ (the case for $\bW$ is similar).
To parallel-update a specific element $(h_j^k)^{(t)}$, we need $(w_k^i)^{(t-1)}$ for all $i \in \Omega_j$,
and $(\bh_j)^{(t-1)}$. We then compute
{\small
\begin{align*}
(a^k_j)_{p}^{(t)} \leftarrow g_1(k,j,p) &:= \sum_{i \in (\Omega_j)_p} \left\{r_j^i+ (w_k^i)^{(t-1)} (h_j^k)^{(t-1)}\right\}(w_t^i)^{(t-1)},  \\
(b^k_j)_{p}^{(t)} \leftarrow g_2(k,j,p) &:= \sum_{i \in (\Omega_j)_p} \left\{(w_k^i)^{(t-1)}\right\}^2,
\end{align*}}
where $\Omega_{j}$ are the (observed) elements of 
column $\bA_j$ in worker $p$'s row-submatrix $\bA^{q_p}$.
Finally, {\bf pull} aggregates the updates:
\begin{equation*}
(h_j^k)^{(t)} \leftarrow g_3(k,j,[(a^k_j)_p^{(t)},(b^k_j)_p^{(t)}]) := \frac{\sum_{p=1}^U {(a^k_j)_{p}^{(t)}}}{\lambda+\sum_{p=1}^U {(b^k_j)_{p}^{(t)}}},
\end{equation*}
with a similar definition for updating $\bW$ using $(w^i_k)^{(t)} \leftarrow f_3()$ and $f_1(i,k,p)$, $f_2(i,k,p)$.
This {\bf push-pull} scheme is free from parallelization error: when $\bW$ are updated by {\bf push}, they
are mutually independent because $\bH$ is held fixed, and vice-versa.
Figure \ref{fig:strads_mf} shows the STRADS MF pseudocode.

\begin{figure}[H] 
{\scriptsize
\begin{Verbatim}[frame=single]
// STRADS Matrix Factorization
\end{Verbatim}
\vspace{-0.3cm}
\begin{Verbatim}[frame=single]
schedule() {
  // Round-robin scheduling
  if counter <= U     // Do W
    return W[q_counter]
  else                // Do H
    return H[r_(counter-U)]
}
\end{Verbatim}
\vspace{-0.3cm}
\begin{Verbatim}[frame=single]
push(worker = p, vars = X[s]) {
  z = []            // Empty list
  if counter <= U   // X is from W
    for row in s, k=1..K
      z.append( (f_1(row,k,p),f_2(row,k,p)) )
  else              // X is from H
    for col in s, k=1..K
      z.append( (g_1(k,col,p),g_2(k,col,p)) )
  return z
}
\end{Verbatim}
\vspace{-0.3cm}
\begin{Verbatim}[frame=single]
pull(workers=[p], vars=X[s], updates=[z]) {
  if counter <= U   // X is from W
    for row in s, k=1..K
      W[row,k] = f_3(row,k,[z])
  else              // X is from H
    for col in s, k=1..K
      H[k,col] = g_3(k,col,[z])
  counter = (counter mod 2*U) + 1
}
\end{Verbatim}
}
\vspace{-0.5cm}
\caption{\small{\bf STRADS MF pseudocode.} Definitions for $f_1,g_1,\dots$
and $q_p,r_p$ are in the text. {\tt counter} is a global model variable.}
\label{fig:strads_mf}
\end{figure}

\subsection{Lasso}
STRADS not only supports simple static {\bf schedules}, but also dynamic, adaptive strategies that
take the model state into consideration.
Consider Lasso regression \cite{tibshirani1996regression}, which
discovers a small subset of features/dimensions that predict the output $\by$.
While Lasso can be solved by random parallelization over each dimension's coefficients,
this strategy fails to converge in the presence of strong dependencies between dimensions~\cite{bradley2011parallel}.
Our STRADS Lasso implementation tackles this challenge by (1) avoiding the simultaneous update of coefficients
whose dimensions are highly inter-dependent, and (2) prioritizing coefficients that contribute the most to algorithm
convergence. These properties complement each other in an algorithmically efficient way, as we shall see.

Formally, Lasso can be defined as an optimization problem:
\\[-0.5cm]
\begin{align}
\min_{\bbeta} \ell(\bX, \by, \bbeta) + \lambda \sum_{j} |\beta_j|,
\label{eq:lasso}
\end{align}\\[-0.3cm]
where $\lambda$ is a regularization parameter that determines the
sparsity of $\bbeta$, and $\ell(\cdot)$ is a non-negative convex loss function such 
as squared-loss or logistic-loss; 
we assume that $\bX$ and $\by$ are standardized and consider \eqref{eq:lasso} without an intercept.
For simplicity but without loss of generality, we let  
$\ell(\bX,\by,\bbeta) = \frac{1}{2}\left\| \by - \bX \bbeta \right\|_2^2$,
and note that it is straightforward to use other loss functions.
Lasso can be solved using coordinate descent (CD) updates \cite{friedman2007pathwise};
by taking the gradient of \eqref{eq:lasso}, we obtain the CD update rule for $\beta_j$:
\\[-0.7cm]
\begin{align}
\beta_j^{(t)} \leftarrow S(
\bx_j^T\by 
-\sum_{k \neq j} \bx_j^T \bx_k \beta_k^{(t-1)},\lambda),
\label{eq:update_rule}
\end{align}\\[-0.3cm]
where $S(\cdot,\lambda)$ is a soft-thresholding operator \cite{friedman2007pathwise}, 
defined by $S(\beta_j,\lambda) \equiv \mbox{sign}(\beta)\left( \left| \beta \right| - \lambda \right)$.

\paragraph{STRADS implementation:}

Our Lasso {\bf schedule} strategy picks variables
dynamically, according to the model state. First, we define
a probability distribution $\mathbf{c}=[c_1,\dots,c_j]$ over the $\bbeta$; the purpose of $\mathbf{c}$
is to prioritize $\beta_j$'s during {\bf schedule}, and thus speed up convergence. In particular,
we observe that choosing $\beta_j$ with probability
$c_j = f_1(j) :\propto |\beta_j^{(t_{j}-2)}-\beta_j^{(t_{j}-1)}| + \eta$
substantially speeds up the Lasso convergence rate (see supplement for our theoretical motivation),
where $\eta$ is a small positive constant, and $t_j$ is the iteration counter for the $j$-th variable.

To prevent non-convergence due to dimension inter-dependencies~\cite{bradley2011parallel},
we only {\bf schedule} $\beta_j$ and $\beta_k$ for concurrent updates 
if $\bx_j^T \bx_k \approx 0$.
This is performed as follows: first, select $U'$ candidates $\beta_j$s 
from the probability distribution $\mathbf{c}$ to form
a set $\mathcal{C}$. Next, choose a subset $\mathcal{B}\subseteq\mathcal{C}$ of size $U\leq U'$
such that $\bx_j^T \bx_k < \rho$ for all $j,k \in \mathcal{B}$, where $\rho \in (0,1]$;
we represent this selection procedure\footnote{
Note that this procedure is inexpensive: by selecting $U'$ candidate $\beta$'s first,
only $U'^2$ dependencies need to be checked, as opposed to $J^2$ where $J$ is the total number of $\beta$.}
by the function $f_2(\mathcal{C})$.
Here $U'$ and $\rho$ are user-defined parameters.
We will show that this {\bf schedule} with sufficiently large $U'$ and small $\rho$
greatly speeds up convergence over naive random scheduling.

Finally, we execute {\bf push} and {\bf pull} to update the $\{\beta_j\}\in\mathcal{B}$ using $U$ workers in parallel.
The rows of the data matrix $\bX$ are partitioned into $U$ submatrices, and the $p$-th worker stores the submatrix $\bX^p$;
With $\bX$ partitioned in this manner, we need to modify the update
rule Eq. \eqref{eq:update_rule} accordingly. Using $U$ workers, {\bf push} computes
$U$ partial summations for each selected $\beta_j$, denoted by 
$\{z_{j,1}^{(t)},\ldots, z_{j,U}^{(t)}\}$, where
$z_{j,p}^{(t)}$ represents the partial summation for the $j$-th $\bbeta$ in the $p$-th worker
at the $t$-th iteration:
\\[-0.4cm]
\begin{align}
z_{j,p}^{(t)} \leftarrow f_3(p,j) :=
(\bx^{p}_{j})^T\by 
-\sum_{k \neq j} ({\bx^{p}_{j}})^T (\bx^{p}_k) \beta_k^{(t-1)}
\label{eq:update_rule_part}
\end{align}\\[-0.3cm]
After all {\bf pushes} have been completed,
{\bf pull} updates $\beta_j$ via
$\beta_j^{(t)} = f_4(j,[z^{(t)}_{j,p}]) := S(\sum_{p=1}^U z_{j,p}^{(t)},\lambda)$.
Figure \ref{fig:strads_lasso} illustrates the STRADS LASSO pseudocode.

\begin{figure}[H] 
{\scriptsize
\begin{Verbatim}[frame=single]
// STRADS Lasso
\end{Verbatim}
\vspace{-0.3cm}
\begin{Verbatim}[frame=single]
schedule() {
  // Priority-based scheduling
  for all j      // Get new priorities
    c_j = f_1(j)
  for a=1..U'    // Prioritize betas
    random draw s_a using [c_1, ..., c_j]
  // Get 'safe' betas
  (j_1, ..., j_U) = f_2(s_1, ..., s_U')
  return (b[j_1], ..., b[j_U])
}
\end{Verbatim}
\vspace{-0.3cm}
\begin{Verbatim}[frame=single]
push(worker = p, vars = (b[j_1],...,b[j_U])) {
  z = []        // Empty list
  for a=1..U    // Compute partial sums
    z.append( f_3(p,j_a) )
  return z
}
\end{Verbatim}
\vspace{-0.3cm}
\begin{Verbatim}[frame=single]
pull(workers = [p], vars = (b[j_1],...,b[j_U]),
     updates = [z]) {
  for a=1..U        // Aggregate partial sums
    b[j_a] = f_4(j_a,[z])
}
\end{Verbatim}
}
\vspace{-0.5cm}
\caption{\small {\bf STRADS Lasso pseudocode.} Definitions for $f_1,f_2,\dots$
are given in the text.}
\label{fig:strads_lasso}
\end{figure}

\section{Experiments}
\label{sec:experiments}

We now demonstrate that our STRADS implementations of LDA, MF and Lasso can
(1) reach larger model sizes than other baselines;
(2) converge at least as fast, if not faster, than other baselines;
(3) with additional machines, STRADS uses less memory per machine (efficient partitioning). 
For baselines, we used (a) a STRADS implementation 
of distributed Lasso with only a naive round-robin scheduler (Lasso-RR), 
(b) GraphLab's Alternating Least Squares (ALS) implementation of MF~\cite{low2012distributed},
(c) YahooLDA for topic modeling~\cite{ahmed2012scalable}.
Note that Lasso-RR imitates the random scheduling scheme proposed by Shotgun
algorithm on STRADS.
We chose GraphLab and YahooLDA, as they are popular choices for distributed MF and LDA.

We conducted experiments on two clusters \cite{gibson2013probe} (with
2-core and 16-core machines respectively),
to show the effectiveness of STRADS model-parallelism across different hardware.
We used the 2-core cluster for LDA, and the 16-core cluster for Lasso and MF.
The 2-core cluster contains 128 machines, each with two 2.6GHz AMD
cores and 8GB RAM, and connected via a 1Gbps network  interface.
The 16-core cluster contains 9 
machines, each with 16 2.1GHz AMD cores and 64GB RAM, and connected via a
40Gbps network interface. All our experiments use a fixed data size, and we
vary the number of machines and/or the model size (unless otherwise stated).

\subsection{Datasets}

\paragraph{Latent Dirichlet Allocation}
We used $3.9$M English Wikipedia abstracts,
and conducted experiments using both unigram (1-word) tokens ($V=2.5$M unique unigrams, $179$M tokens)
and bigram (2-word) tokens ($V=21.8$M unique bigrams, $79$M tokens).
We note that our bigram vocabulary ($21.8$M) is an order of magnitude larger than
recently published results~\cite{ahmed2012scalable},
demonstrating that STRADS scales to very large models.
We set the number of topics to $K=5000$ and $10000$ (again, significantly
larger than recent literature~\cite{ahmed2012scalable}), which creates
extremely large word-topic tables: $12.5$B elements (unigram) and $109$B elements (bigram).

\paragraph{Matrix Factorization}
We used the Nexflix dataset \cite{bennett2007netflix} for our MF experiments:
100M anonimized ratings from 480,189 users on 17,770 movies.
We varied the rank of $\bW,\bH$ from $K=20$ to $2000$, which
exceeds the upper limit of previous MF papers~\cite{zhou2008large,gemulla2011large,yu2012scalable}.

\paragraph{Lasso}
We used synthetic data with 50K samples and $J=10$M to 100M features, 
where every feature $\bx_j$ has only 25 non-zero samples.
To simulate correlations between adjacent features (which exist in real-world data),
we first added $Unif(0,1)$ noise to $\bx_1$.
Then, for $j=2, \ldots, J$, with 0.9 probability we add $\epsilon_j = Unif(0,1)$ noise to $\bx_j$,
otherwise we add $0.9 \epsilon_{j-1} + 0.1 Unif(0,1)$ to $\bx_j$.

\subsection{Speed and Model Sizes}

\begin{figure*}[ht]
\centering
\includegraphics[width=0.3\textwidth]{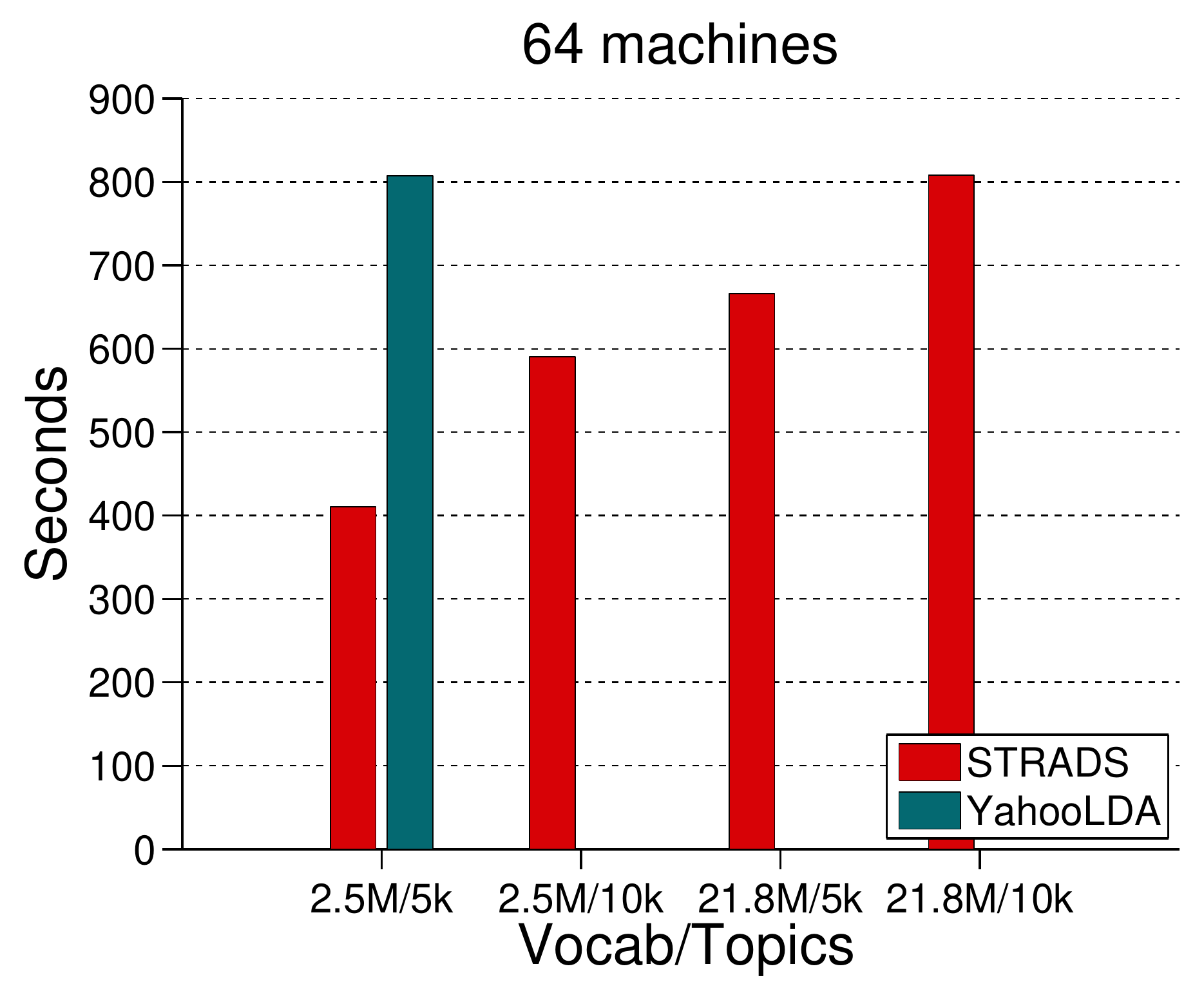}
\includegraphics[width=0.3\textwidth]{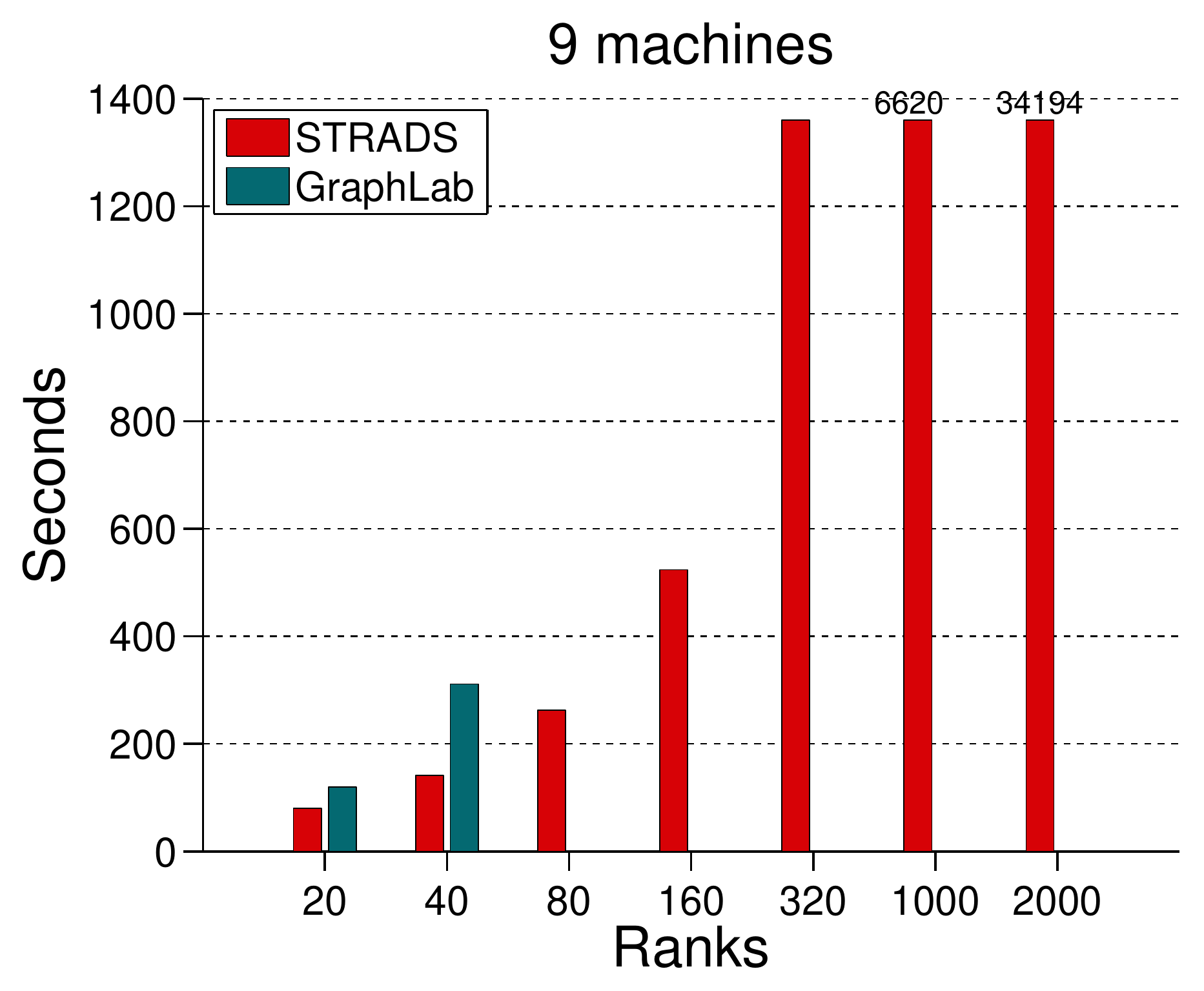}
\includegraphics[width=0.3\textwidth]{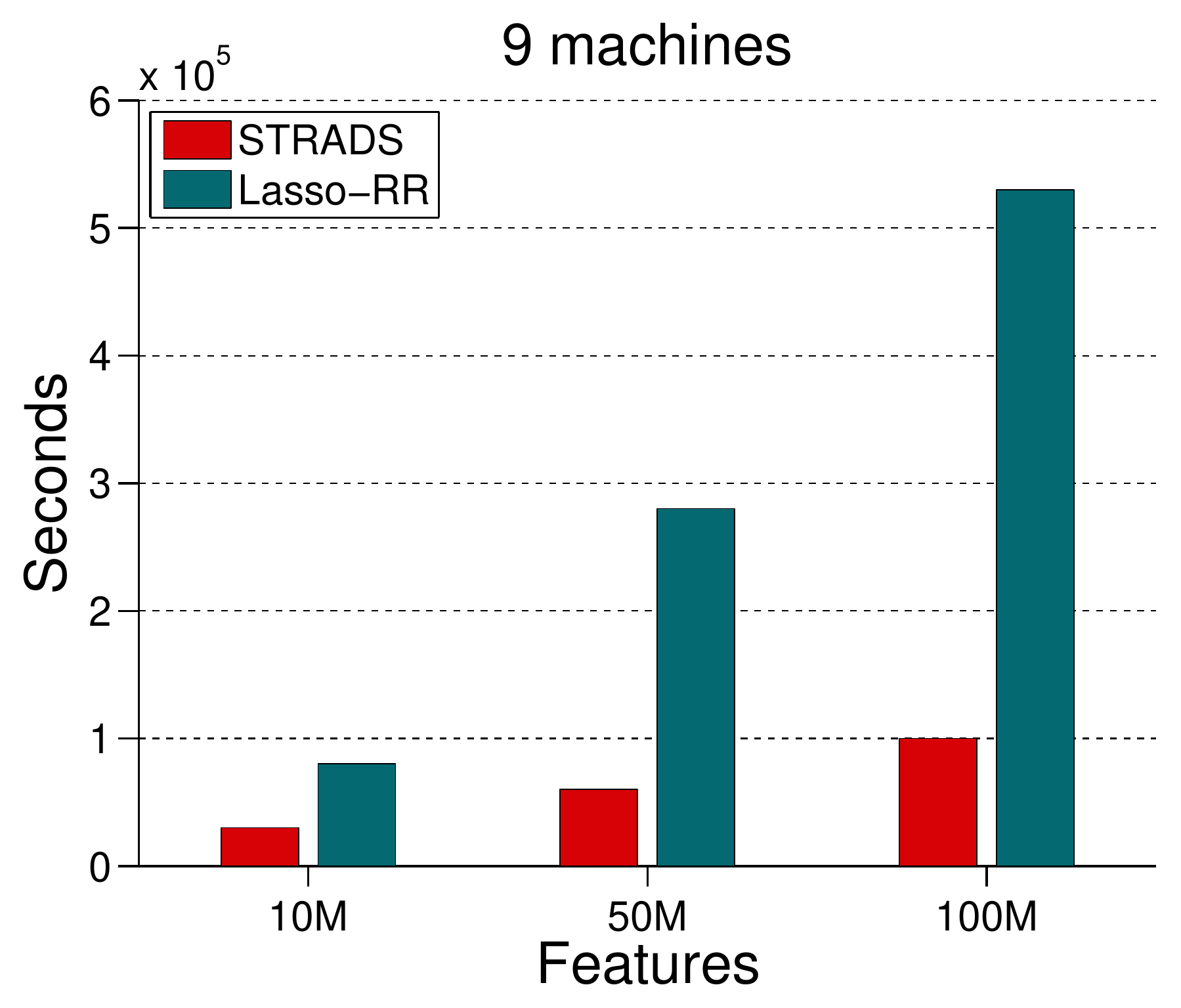}
\caption{\small Convergence time
versus model size for STRADS and baselines
for (left) LDA, (center) MF, and (right) Lasso.
We omit the bars if a method did not reach 98\% of STRADS's convergence point
(YahooLDA
and GraphLab-MF  failed at 2.5M-Vocab/10K-topics and rank $K\geq 80$, respectively). 
STRADS not only reaches larger model sizes than YahooLDA, GraphLab, and Lasso-RR,  
but also converges significantly faster.}
\label{fig:convergencetime_vs_K}
\end{figure*}
\begin{figure}[ht]
\centering
\includegraphics[width=0.3\textwidth]{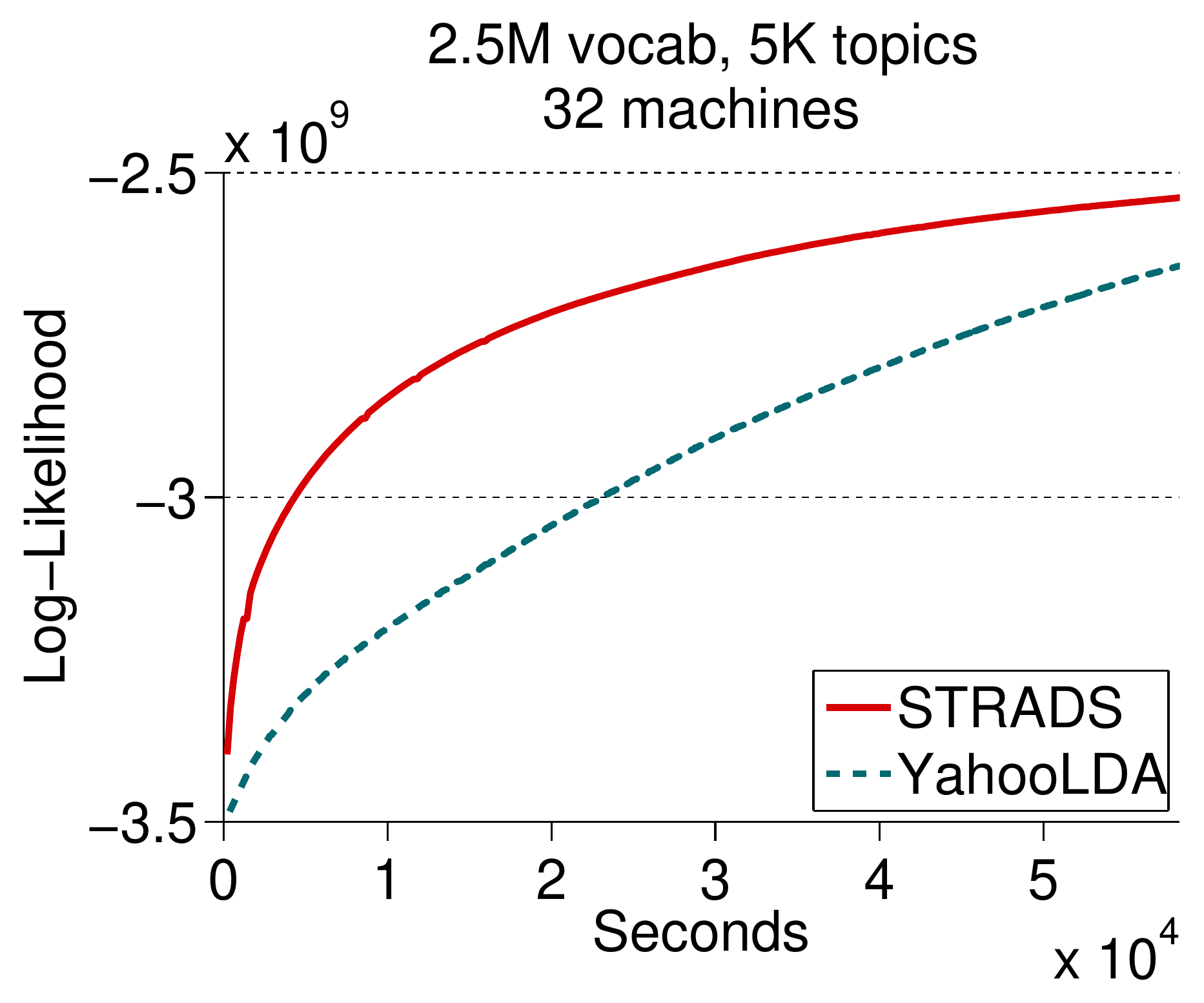}
\includegraphics[width=0.3\textwidth]{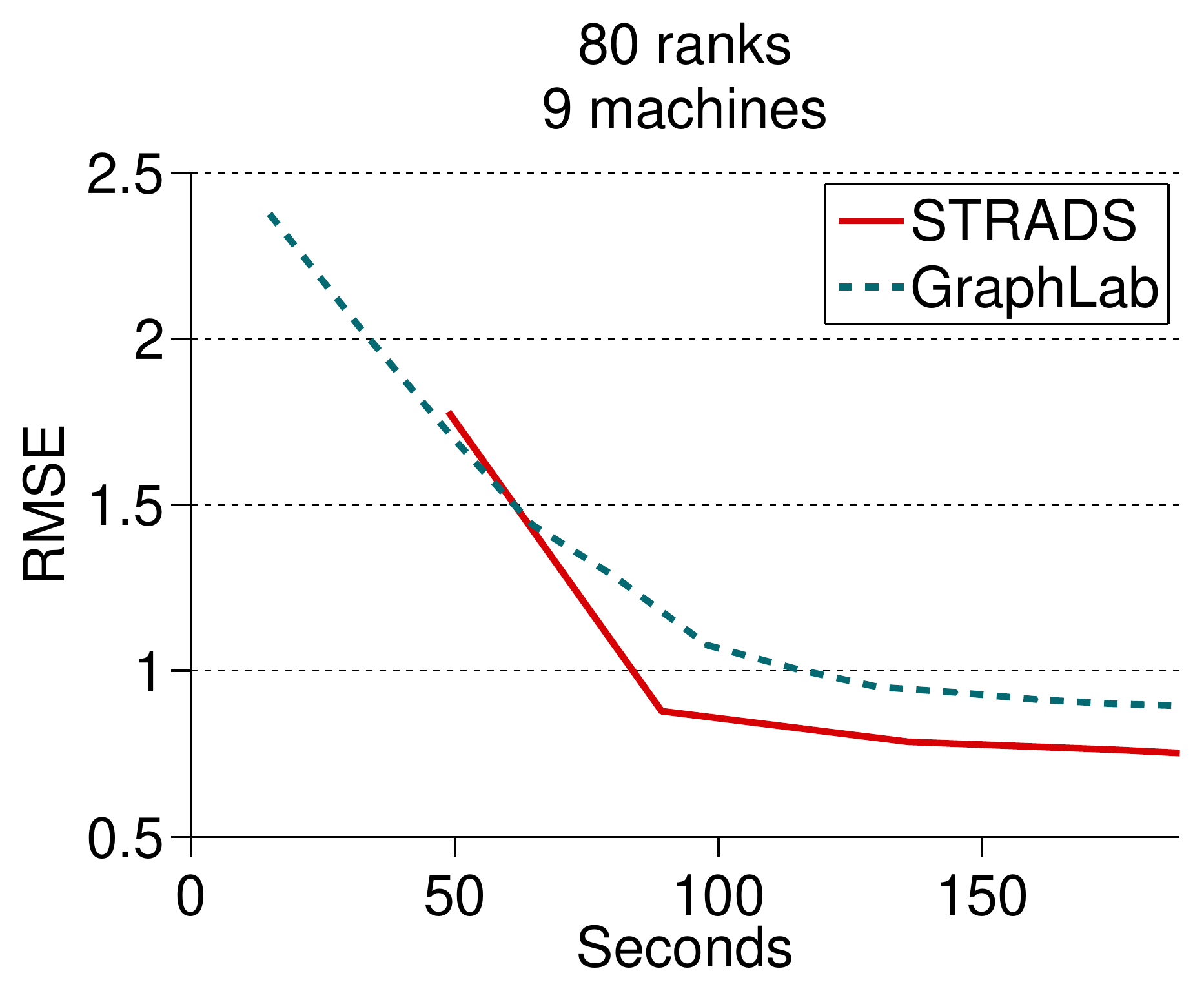}
\includegraphics[width=0.3\textwidth]{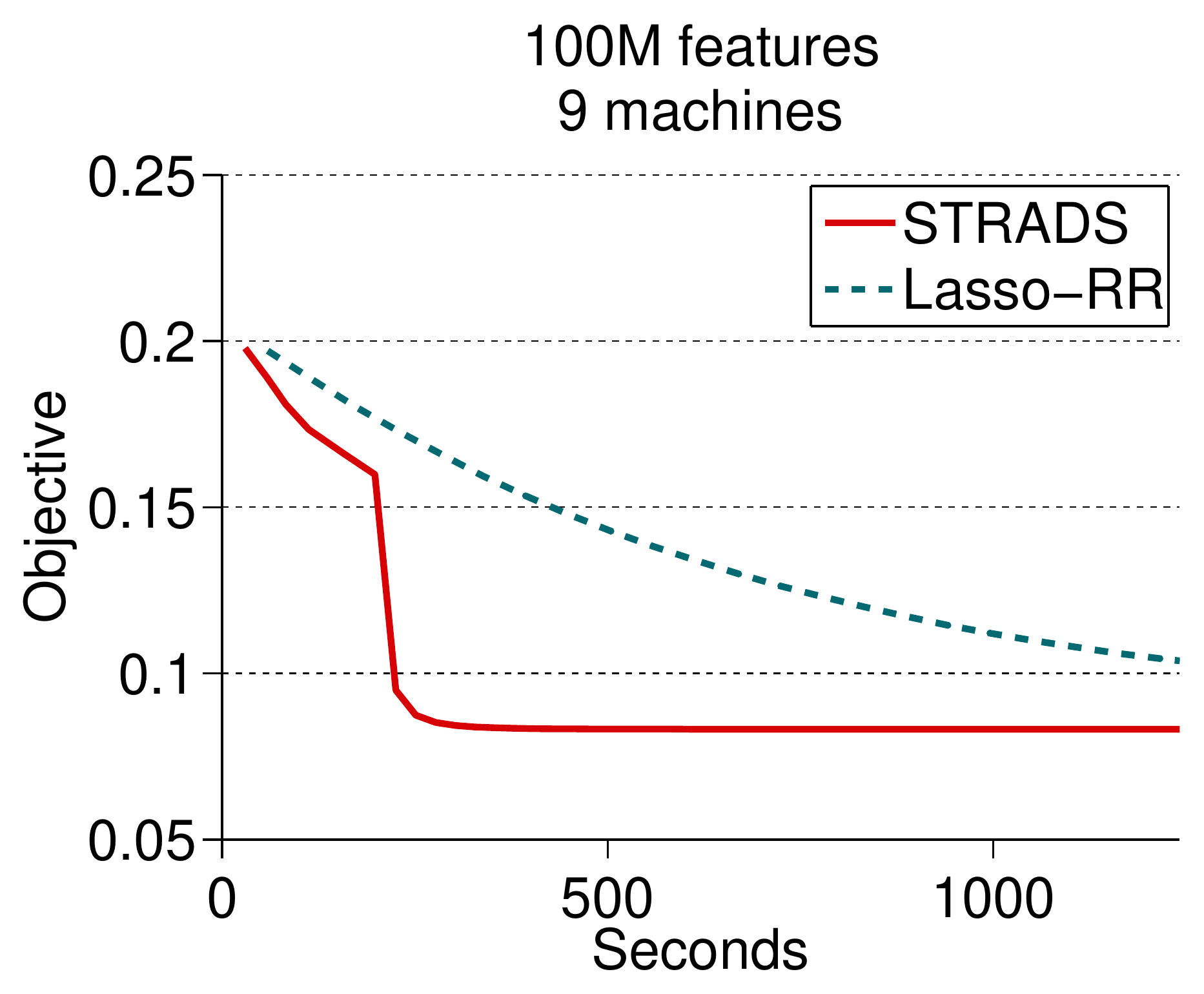}
\vspace{-0.5cm}
\caption{\small Convergence trajectories of different methods
for (left) LDA, (center) MF, and (right) Lasso.}
\label{fig:convergence}
\end{figure}

Figure \ref{fig:convergencetime_vs_K} shows the time taken by each algorithm
to reach a fixed objective value (over a range of model sizes), as well as
the largest model size that each baseline was capable of running.
For LDA and MF, STRADS handles much larger model sizes than either YahooLDA (could only handle 5K topics
on the unigram dataset) or GraphLab (could only handle rank $<80$),
while converging more quickly; we attribute STRADS's faster convergence
to lower parallelization error (LDA only) and reduced synchronization requirements
through careful model partitioning (LDA, MF).
In particular, YahooLDA stores nearly the whole word-topic table on every machine,
so its maximum model size is limited by the smallest machine (Figure~\ref{fig:lda_memoryusage_permachine}).
For Lasso, STRADS converges more quickly than Lasso-RR because of our dynamic {\bf schedule} strategy,
which is graphically captured in the convergence trajectory seen in Figure \ref{fig:convergence} ---
observe that STRADS's dynamic {\bf schedule} causes the Lasso objective to plunge quickly to the optimum
at around 250 seconds. We also see that STRADS LDA and MF achieved better objective values,
confirming that STRADS model-parallelism is fast without compromising convergence quality.

\subsection{Scalability}

\begin{figure}[t] 
\centering
\includegraphics[width=0.35\textwidth]{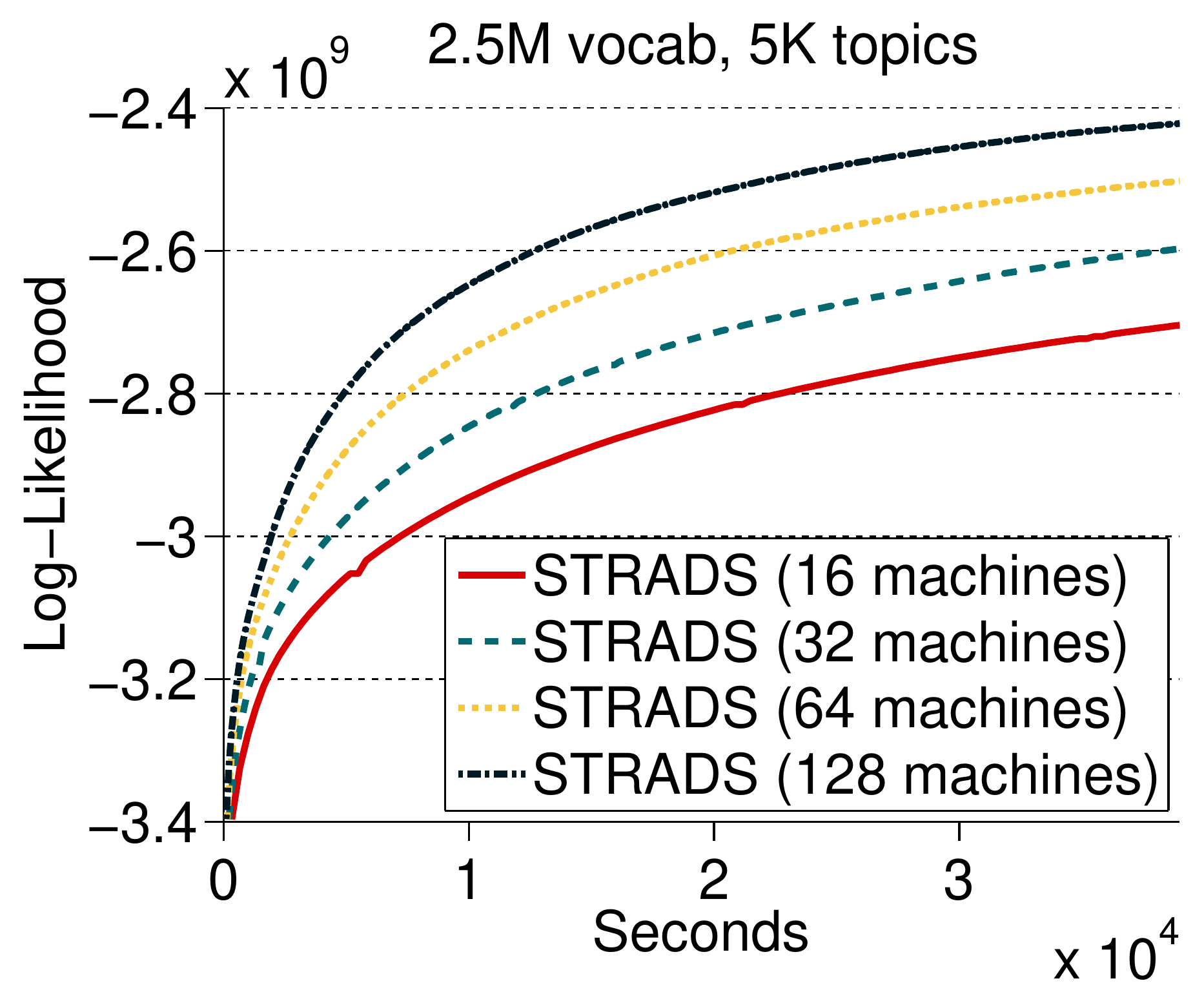}
\includegraphics[width=0.35\textwidth]{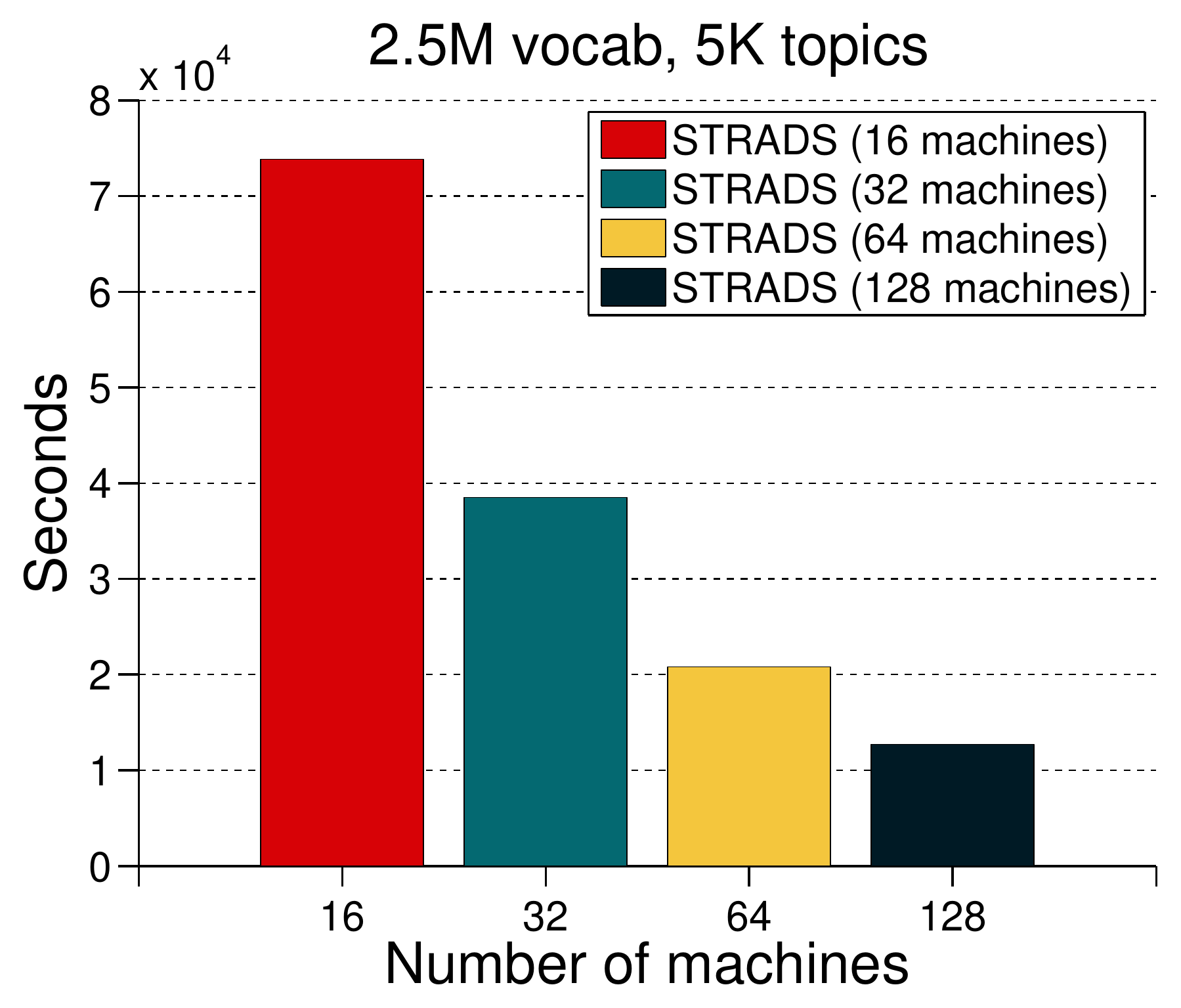}
\caption{\small STRADS LDA scalablity with increasing machines
using a fixed model size. (Left) Convergence trajectories;
(Right) Time taken to reach a log-likelihood of
$-2.6 \times 10^9$.
}
\label{fig:lda_conv_machine}
\end{figure}

In Figure \ref{fig:lda_conv_machine}, we show the convergence trajectories and time-to-convergence
for STRADS LDA using different numbers of machines at a fixed model size (unigram with 2.5M vocab and 5K topics).
The plots confirm that STRADS LDA exhibits faster convergence with more machines, and that
the time to convergence almost halves with every doubling of machines (near-linear scaling).

\section{Discussion and Related Work}

As a framework of user-programmable primitives for dynamic
Big Model-parallelism, STRADS provides the following benefits:
(1) scalability and efficient memory utilization,
allowing larger models to be run with additional machines
(because the model is partitioned, rather than duplicated across machines);
(2) the ability to invoke dynamic {\bf schedules} that reduce model variable dependencies across
workers, leading to lower parallelization error and thus faster, correct convergence.

While the notion of model-parallelism is not new, our contribution is to study
it within the context of a programmable system (STRADS), using primitives that enable
general, user-programmable partitioning and static/dynamic scheduling of variable updates (based on model
dependencies).
Previous works explore aspects of model-parallelism in a more specific context:
Scherrer et al.~\cite{scherrer2012feature} proposed a
static model partitioning scheme specifically for parallel coordinate descent,
while GraphLab~\cite{Low+al:uai10graphlab,low2012distributed} 
statically pre-partitions data and variables through a graph abstraction.

An important direction for future research is to reduce the communication costs of using
STRADS. Currently, STRADS adopts a star topology from scheduler machines to workers,
which causes the scheduler to eventually become a bottleneck as we increase the number of machines.
To mitigate this issue, we wish to explore different {\bf sync} schemes
such as an asynchronous parallelism~\cite{ahmed2012scalable}
and stale synchronous parallelism~\cite{ho2013more}.
We also want to explore the use of STRADS for other popular ML applications,
such as support vector machines and logistic regression.

{\small
\bibliographystyle{plain}

\begin{thebibliography}{10}

\bibitem{ahmed2012scalable}
Amr Ahmed, Moahmed Aly, Joseph Gonzalez, Shravan Narayanamurthy, and
  Alexander~J Smola.
\newblock Scalable inference in latent variable models.
\newblock In {\em WSDM}, pages 123--132. ACM, 2012.

\bibitem{bennett2007netflix}
James Bennett and Stan Lanning.
\newblock The netflix prize.
\newblock In {\em Proceedings of KDD cup and workshop}, volume 2007, page~35,
  2007.

\bibitem{blei2003latent}
David~M Blei, Andrew~Y Ng, and Michael~I Jordan.
\newblock Latent dirichlet allocation.
\newblock {\em the Journal of machine Learning research}, 3:993--1022, 2003.

\bibitem{bradley2011parallel}
Joseph~K Bradley, Aapo Kyrola, Danny Bickson, and Carlos Guestrin.
\newblock Parallel coordinate descent for l1-regularized loss minimization.
\newblock {\em ICML}, 2011.

\bibitem{dean2012large}
J.~Dean, G.~Corrado, R.~Monga, K.~Chen, M.~Devin, Q.~V. Le, M.~Z. Mao,
  M.~Ranzato, A.~W. Senior, P.~A. Tucker, et~al.
\newblock Large scale distributed deep networks.
\newblock In {\em NIPS}, pages 1232--1240, 2012.

\bibitem{dean2008mapreduce}
Jeffrey Dean and Sanjay Ghemawat.
\newblock Map{R}educe: simplified data processing on large clusters.
\newblock {\em Communications of the ACM}, 51(1):107--113, 2008.

\bibitem{fan2009ultrahigh}
Jianqing Fan, Richard Samworth, and Yichao Wu.
\newblock Ultrahigh dimensional feature selection: beyond the linear model.
\newblock {\em The Journal of Machine Learning Research}, 10:2013--2038, 2009.

\bibitem{friedman2007pathwise}
J.~Friedman, T.~Hastie, H.~Hofling, and R.~Tibshirani.
\newblock {Pathwise coordinate optimization}.
\newblock {\em Annals of Applied Statistics}, 1(2):302--332, 2007.

\bibitem{gemulla2011large}
Rainer Gemulla, Erik Nijkamp, Peter~J Haas, and Yannis Sismanis.
\newblock Large-scale matrix factorization with distributed stochastic gradient
  descent.
\newblock In {\em Proceedings of the 17th ACM SIGKDD international conference
  on Knowledge discovery and data mining}, pages 69--77. ACM, 2011.

\bibitem{gibson2013probe}
Garth Gibson, Gary Grider, Andree Jacobson, and Wyatt Lloyd.
\newblock Probe: A thousand-node experimental cluster for computer systems
  research.
\newblock {\em USENIX; login}, 38, 2013.

\bibitem{gonzalez2011parallel}
J.~Gonzalez, Y.~Low, A.~Gretton, and C.~Guestrin.
\newblock Parallel gibbs sampling: From colored fields to thin junction trees.
\newblock In {\em International Conference on Artificial Intelligence and
  Statistics}, pages 324--332, 2011.

\bibitem{griffiths2004finding}
Thomas~L Griffiths and Mark Steyvers.
\newblock Finding scientific topics.
\newblock {\em Proceedings of the National Academy of Sciences of the United
  States of America}, 101(Suppl 1):5228--5235, 2004.

\bibitem{ho2013more}
Q.~Ho, J.~Cipar, H.~Cui, J.-K. Kim, S.~Lee, P.~B. Gibbons, G.~Gibson, G.~R.
  Ganger, and E.~P. Xing.
\newblock More effective distributed ml via a stale synchronous parallel
  parameter server.
\newblock In {\em NIPS}, 2013.

\bibitem{low2012distributed}
Y.~Low, J.~Gonzalez, A.~Kyrola, D.~Bickson, C.~Guestrin, and J.~M. Hellerstein.
\newblock {Distributed GraphLab: A Framework for Machine Learning and Data
  Mining in the Cloud}.
\newblock {\em PVLDB}, 2012.

\bibitem{Low+al:uai10graphlab}
Yucheng Low, Joseph Gonzalez, Aapo Kyrola, Danny Bickson, Carlos Guestrin, and
  Joseph~M. Hellerstein.
\newblock Graphlab: A new parallel framework for machine learning.
\newblock In {\em UAI}, July 2010.

\bibitem{malewicz2010pregel}
Grzegorz Malewicz, Matthew~H Austern, Aart~JC Bik, James~C Dehnert, Ilan Horn,
  Naty Leiser, and Grzegorz Czajkowski.
\newblock Pregel: a system for large-scale graph processing.
\newblock In {\em Proceedings of the 2010 ACM SIGMOD International Conference
  on Management of data}, pages 135--146. ACM, 2010.

\bibitem{newman2009distributed}
David Newman, Arthur Asuncion, Padhraic Smyth, and Max Welling.
\newblock Distributed algorithms for topic models.
\newblock {\em The Journal of Machine Learning Research}, 10:1801--1828, 2009.

\bibitem{scherrer2012feature}
Chad Scherrer, Ambuj Tewari, Mahantesh Halappanavar, and David Haglin.
\newblock Feature clustering for accelerating parallel coordinate descent.
\newblock {\em NIPS}, 2012.

\bibitem{tibshirani1996regression}
R.~Tibshirani.
\newblock {Regression shrinkage and selection via the lasso}.
\newblock {\em Journal of the Royal Statistical Society. Series B
  (Methodological)}, 58(1):267--288, 1996.

\bibitem{yao2009efficient}
Limin Yao, David Mimno, and Andrew McCallum.
\newblock Efficient methods for topic model inference on streaming document
  collections.
\newblock 2009.

\bibitem{yu2012scalable}
Hsiang-Fu Yu, Cho-Jui Hsieh, Si~Si, and Inderjit Dhillon.
\newblock Scalable coordinate descent approaches to parallel matrix
  factorization for recommender systems.
\newblock In {\em ICDM 2012}, pages 765--774. IEEE, 2012.

\bibitem{zaharia2010spark}
M.~Zaharia, M.~Chowdhury, M.~J. Franklin, S.~Shenker, and I.~Stoica.
\newblock Spark: cluster computing with working sets.
\newblock In {\em Proceedings of the 2nd USENIX conference on Hot topics in
  cloud computing}, 2010.

\bibitem{zhou2008large}
Y.~Zhou, D.~Wilkinson, R.~Schreiber, and R.~Pan.
\newblock Large-scale parallel collaborative filtering for the netflix prize.
\newblock In {\em Algorithmic Aspects in Information and Management}, pages
  337--348. Springer, 2008.

\bibitem{zinkevich2009slow}
Martin Zinkevich, John Langford, and Alex~J Smola.
\newblock Slow learners are fast.
\newblock In {\em Advances in Neural Information Processing Systems}, pages
  2331--2339, 2009.

\bibitem{zinkevich2010parallelized}
Martin Zinkevich, Markus Weimer, Lihong Li, and Alex~J Smola.
\newblock Parallelized stochastic gradient descent.
\newblock In {\em Advances in Neural Information Processing Systems}, pages
  2595--2603, 2010.

\end{thebibliography}

}

\end{document}